\documentclass[sigconf]{acmart}

\AtBeginDocument{%
  \providecommand\BibTeX{{%
    \normalfont B\kern-0.5em{\scshape i\kern-0.25em b}\kern-0.8em\TeX}}}

\setcopyright{acmcopyright}
\copyrightyear{2018}
\acmYear{2018}
\acmDOI{XXXXXXX.XXXXXXX}

\acmJournal{TOG}
\acmVolume{37}
\acmNumber{4}
\acmArticle{111}
\acmMonth{8}

\begin{document}

\title{
Communicating Robot's Intentions while Assisting Users via Augmented Reality}

\author{Chao Wang}
\authornote{Equal contribution. Listing order is random. 
Chao designed and implemented the AR-based user interface. Theodoros and Micheal proposed the ergonomics model and implemented the robotic system. Anna contributed to the UX design and the scenario generation.}
\email{chao.wang@honda-ri.de}
\affiliation{%
  \institution{Honda Research Institute Europe}
  \city{Offenbach}
  \state{Hessen}
  \country{Germany}
}

\author{Theodoros Stouraitis}
\authornotemark[1]
\email{theodoros.stouraitis@honda-ri.de}
\affiliation{
  \institution{Honda Research Institute Europe}
  \city{Offenbach}
  \state{Hessen}
  \country{Germany}
}

\author{Anna Belardinelli}
\authornotemark[1]
\email{anna.belardinelli@honda-ri.de}
\affiliation{
  \institution{Honda Research Institute Europe}
  \city{Offenbach}
  \state{Hessen}
  \country{Germany}
}

\author{Stephan Hasler}
\email{stephan.hasler@honda-ri.de}
\affiliation{
  \institution{Honda Research Institute Europe}
  \city{Offenbach}
  \state{Hessen}
  \country{Germany}
}

\author{Michael Gienger}
\authornotemark[1]
\email{michael.gienger@honda-ri.de}
\affiliation{
  \institution{Honda Research Institute Europe}
  \city{Offenbach}
  \state{Hessen}
  \country{Germany}
}

\renewcommand{\shortauthors}{Trovato and Tobin, et al.}

\begin{abstract}

This paper explores the challenges faced by assistive robots in effectively cooperating with humans, requiring them to anticipate human behavior, predict their actions' impact, and generate understandable robot actions. The study focuses on a use-case involving a user with limited mobility needing assistance with pouring a beverage, where tasks like unscrewing a cap or reaching for objects demand coordinated support from the robot. Yet, anticipating the robot's intentions can be challenging for the user, which can hinder effective collaboration. To address this issue, we propose an innovative solution that utilizes Augmented Reality (AR) to communicate the robot's intentions and expected movements to the user, fostering a seamless and intuitive interaction.
\end{abstract}

\ccsdesc[500]{Human-centered computing~Interaction techniques}

\keywords{human-robot interaction, augmented reality, collaborative intelligence, human-centered AI}

\begin{teaserfigure}
  \includegraphics[width=\textwidth]{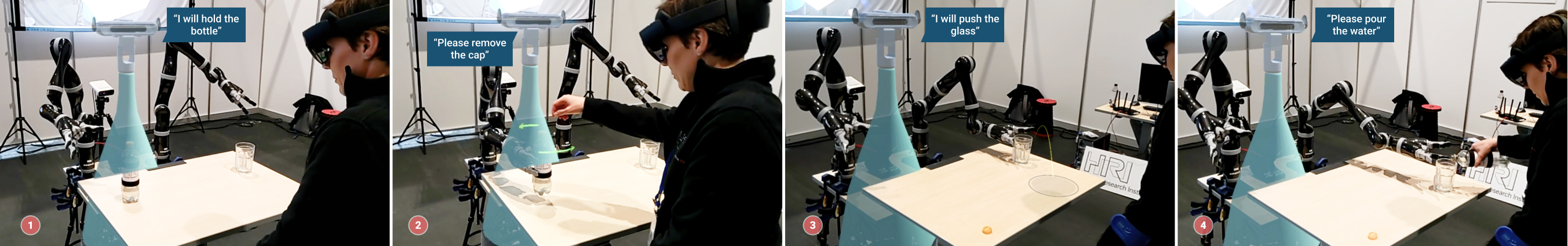}
  \caption{Interaction Flow for assistance in a pouring task}
  \label{fig:teaser}
\end{teaserfigure}

\received{20 February 2007}
\received[revised]{12 March 2009}
\received[accepted]{5 June 2009}

\maketitle

\section{Introduction}
Assistive robots can play a crucial role in the life of more than 200 million humans that need assistance with activities of daily living (ADLs)~\cite{Who2011}, such as getting out of bed, going to the restroom, preparing a meal. 
Efficient cooperation between a human and a robot includes a number of  challenges. These, among others, are: (i) human behaviour anticipation, 
 (ii) predicting the influence of the robot's actions onto the human's actions~\cite{trautman2010unfreezing}, 
and (iii) generating legible robot cues that can be easily understood by the human partner, such as actions~\cite{Dragan_2013}.
Here, we study these challenges in the context of manipulation where a robot assists a human to perform a sequence of table-top manipulation actions. 
In such a scenario, the robot needs to predict a sequence of possible human's actions (discrete decisions) and their outcome (continuous states)~\cite{StouraitisTRO2020}, predict and assess the possible human postures~\cite{kim2017anticipatory}, compute the desired intervention that improves the physical state of the human, and communicate the effects of the intended interventions to the human ahead of time. \par
To achieve these aims, we propose a model-based optimisation approach that performs a physics-based prediction of the human's actions as the primary objective and treats the robot assistance as a secondary objective.
This allows us to investigate the effects of possible (robot) interventions without altering the prediction of the actual human actions, rather only the way these actions are performed, \textit{e.g.} bending over or not. In this way, the robot can decide which support intervention will improve the human's state to be more ergonomic. For seamlessly collaboration, the intention and guidance of the robot are communicated by holograms and speech output via wearable Augmented Reality headset. 
\section{System and Interaction}
\begin{figure}
    \includegraphics[width=0.45\textwidth]{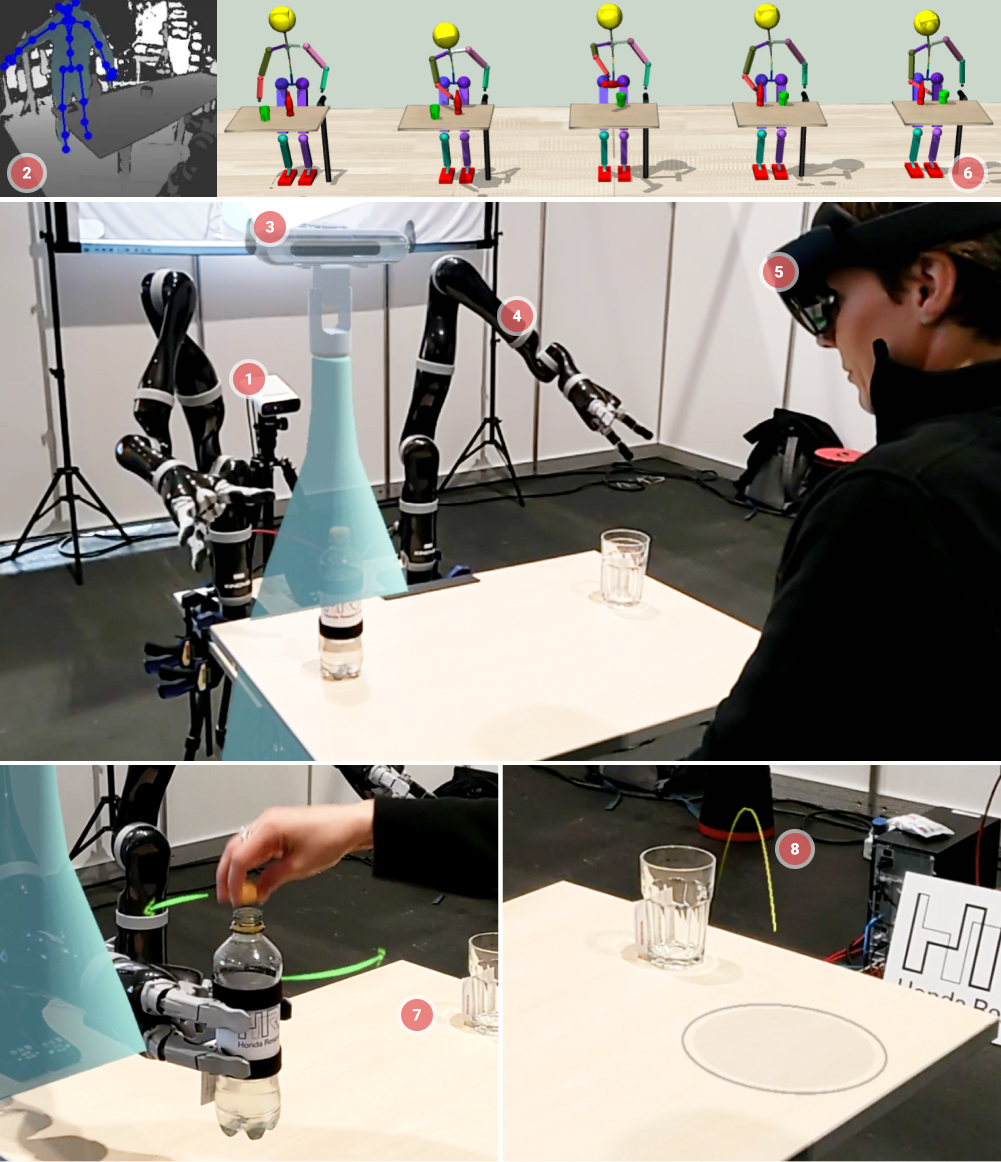}
    \caption{system and interaction design}
    \label{fig:system}
    \centering
\end{figure}
The robot assistance system was developed based on the existing technical framework proposed in~\cite{wang2023explainable}, enabling seamless communication between a robot and a wearable AR headset (MS HoloLens). Our 
implementation involved a mixed reality robot with two real arms and a virtual holographic body, as illustrated in Figure \ref{fig:system} (components 3 and 4). To accurately detect the human posture (2 in Figure \ref{fig:system}) and each object's pose on the table, we integrated 3d body tracking \footnote{Microsoft Azure Kinect DK, https://azure.microsoft.com/en-us/products/kinect-dk} and fiducial marker tracking \cite{garrido2014automatic} into our setup. 
\par
Here, we explain the robot assistance with a "pouring water" example.
The goal of this assistive task is communicated to the robot by speech input. The user can trigger the robot's assistance function by saying "Help me to get some water." The semantic-level steps to achieving this goal are assumed to be known, and are stored in the system's memory. The steps are: 1) grasping and lifting the bottle, 2) unscrewing the bottle cap, 3) putting the cap down, 4) pouring the water into the glass, and 5) picking up the glass to drink. Each of the steps can be seen as a "key-frame", in which the related objects' pose as well as the pose of the actuators (e.g. hand) for manipulating the objects are constrained (6 in Figure \ref{fig:system}). For example, the step of "pouring the water into the glass" requires that the opened bottle neck is centered above the glass and that one hand holds the bottle. The person to be assisted is assumed to have an impairment, which in this example is a left arm leaning on a crutch. The assistance concept is based on computing an optimal arrangement of the entities (objects) of the scene, which is the location of the bottle and of the glass. We employ an optimization formalism that accounts for the potential impairment, the constraints of the above-mentioned key-frames, and computes an object arrangement that leads to comfortable human postures according to a minimum-torque criterion. 
The object poses are then realized by the robot, using a set of atomic actions to grasp, hold, pour, put down, and push the respective objects.
This information is also communicated to the human user with the AR headset. Here we go through each steps of the collaborative task and describe the robot's actions as well as its communication cues with the user: \par
\textbf{1) Grasping and lifting the bottle:} As the user is assumed to have an immobilized left arm, and the bottle is positioned to its left, the robot takes over this step. It uses its speech channel (from the HoloLens) to say: "I will hold the bottle"(1 in Figure \ref{fig:teaser}). Then it picks up the bottle and moves it to a pose that is easy for the human to reach by his right arm. \par
\textbf{2) Unscrewing the cap:} After the bottle is held in front of the user, the robot guides the user by speech: "Please remove the cap"(2 in Figure \ref{fig:teaser}). Meanwhile, two spinning arrows are shown upon the location of the cap (7 in Figure \ref{fig:system}).\par
\textbf{3) Putting the cap and the bottle down:} The robot will ask the human user to put the cap down by speech: "Please put the cap on the table" and it will place the bottle on the table.
\par
\textbf{4) Pouring the water into the glass:} The pouring key-frame requires the bottle and the glasses to be aligned. Considering that the glass is far away from the user and the bottle is close to the user, the robot will firstly ask user to pick up the bottle by speech, then push the glass to the optimal location for the user to pour without uncomfortably leaning on the table, with the speech output: "I will push the glass."(3 in Figure \ref{fig:teaser}) For visualization, a plate-like graphic is shown to indicate the final position of the cup on the table, and a comet-style animation starting from the current position of the glass to the final position is looped every 0.7 second (8 in Figure \ref{fig:system}). Afterwards, the robot will tell the user: "You can pour into the glass." In the end, the user can picks up the glass to drink the water.
\section{Conclusion and Future works}
This system was tested with around 100 researchers and naive users at different exhibitions which validated its effectiveness and seamless interaction. Future work includes exploring additional use cases and conducting in-depth user studies to advance this concept further, leading to more user-friendly and practical assistive robot systems for enhanced human-robot cooperation.

\bibliographystyle{ACM-Reference-Format}
\bibliography{ar-ergo}


\begin{thebibliography}{7}


\ifx \showCODEN    \undefined \def \showCODEN     #1{\unskip}     \fi
\ifx \showDOI      \undefined \def \showDOI       #1{#1}\fi
\ifx \showISBNx    \undefined \def \showISBNx     #1{\unskip}     \fi
\ifx \showISBNxiii \undefined \def \showISBNxiii  #1{\unskip}     \fi
\ifx \showISSN     \undefined \def \showISSN      #1{\unskip}     \fi
\ifx \showLCCN     \undefined \def \showLCCN      #1{\unskip}     \fi
\ifx \shownote     \undefined \def \shownote      #1{#1}          \fi
\ifx \showarticletitle \undefined \def \showarticletitle #1{#1}   \fi
\ifx \showURL      \undefined \def \showURL       {\relax}        \fi
\providecommand\bibfield[2]{#2}
\providecommand\bibinfo[2]{#2}
\providecommand\natexlab[1]{#1}
\providecommand\showeprint[2][]{arXiv:#2}

\bibitem[Dragan et~al\mbox{.}(2013)]%
        {Dragan_2013}
\bibfield{author}{\bibinfo{person}{Anca~D. Dragan},
  \bibinfo{person}{Kenton~C.T. Lee}, {and} \bibinfo{person}{Siddhartha~S.
  Srinivasa}.} \bibinfo{year}{2013}\natexlab{}.
\newblock \showarticletitle{Legibility and predictability of robot motion}. In
  \bibinfo{booktitle}{\emph{2013 8th ACM/IEEE International Conference on
  Human-Robot Interaction (HRI)}}. \bibinfo{pages}{301--308}.
\newblock
\urldef\tempurl%
\url{https://doi.org/10.1109/HRI.2013.6483603}
\showDOI{\tempurl}


\bibitem[Garrido-Jurado et~al\mbox{.}(2014)]%
        {garrido2014automatic}
\bibfield{author}{\bibinfo{person}{Sergio Garrido-Jurado},
  \bibinfo{person}{Rafael Mu{\~n}oz-Salinas},
  \bibinfo{person}{Francisco~Jos{\'e} Madrid-Cuevas}, {and}
  \bibinfo{person}{Manuel~Jes{\'u}s Mar{\'\i}n-Jim{\'e}nez}.}
  \bibinfo{year}{2014}\natexlab{}.
\newblock \showarticletitle{Automatic generation and detection of highly
  reliable fiducial markers under occlusion}.
\newblock \bibinfo{journal}{\emph{Pattern Recognition}} \bibinfo{volume}{47},
  \bibinfo{number}{6} (\bibinfo{year}{2014}), \bibinfo{pages}{2280--2292}.
\newblock


\bibitem[Kim et~al\mbox{.}(2017)]%
        {kim2017anticipatory}
\bibfield{author}{\bibinfo{person}{Wansoo Kim}, \bibinfo{person}{Jinoh Lee},
  \bibinfo{person}{Luka Peternel}, \bibinfo{person}{Nikos Tsagarakis}, {and}
  \bibinfo{person}{Arash Ajoudani}.} \bibinfo{year}{2017}\natexlab{}.
\newblock \showarticletitle{Anticipatory robot assistance for the prevention of
  human static joint overloading in human--robot collaboration}.
\newblock \bibinfo{journal}{\emph{IEEE robotics and automation letters}}
  \bibinfo{volume}{3}, \bibinfo{number}{1} (\bibinfo{year}{2017}),
  \bibinfo{pages}{68--75}.
\newblock


\bibitem[Stouraitis et~al\mbox{.}(2020)]%
        {StouraitisTRO2020}
\bibfield{author}{\bibinfo{person}{Theodoros Stouraitis},
  \bibinfo{person}{Iordanis Chatzinikolaidis}, \bibinfo{person}{Michael
  Gienger}, {and} \bibinfo{person}{Sethu Vijayakumar}.}
  \bibinfo{year}{2020}\natexlab{}.
\newblock \showarticletitle{Online Hybrid Motion Planning for Dyadic
  Collaborative Manipulation via Bilevel Optimization}.
\newblock \bibinfo{journal}{\emph{IEEE Transactions on Robotics}}
  \bibinfo{volume}{36}, \bibinfo{number}{5} (\bibinfo{year}{2020}),
  \bibinfo{pages}{1452--1471}.
\newblock
\urldef\tempurl%
\url{https://doi.org/10.1109/TRO.2020.2992987}
\showDOI{\tempurl}


\bibitem[Trautman and Krause(2010)]%
        {trautman2010unfreezing}
\bibfield{author}{\bibinfo{person}{Peter Trautman} {and}
  \bibinfo{person}{Andreas Krause}.} \bibinfo{year}{2010}\natexlab{}.
\newblock \showarticletitle{Unfreezing the robot: Navigation in dense,
  interacting crowds}. In \bibinfo{booktitle}{\emph{2010 IEEE/RSJ International
  Conference on Intelligent Robots and Systems}}. IEEE,
  \bibinfo{pages}{797--803}.
\newblock


\bibitem[Wang et~al\mbox{.}(2023)]%
        {wang2023explainable}
\bibfield{author}{\bibinfo{person}{Chao Wang}, \bibinfo{person}{Anna
  Belardinelli}, \bibinfo{person}{Stephan Hasler}, \bibinfo{person}{Theodoros
  Stouraitis}, \bibinfo{person}{Daniel Tanneberg}, {and}
  \bibinfo{person}{Michael Gienger}.} \bibinfo{year}{2023}\natexlab{}.
\newblock \showarticletitle{Explainable Human-Robot Training and Cooperation
  with Augmented Reality}. In \bibinfo{booktitle}{\emph{Extended Abstracts of
  the 2023 CHI Conference on Human Factors in Computing Systems}}.
  \bibinfo{pages}{1--5}.
\newblock


\bibitem[{World Health Organization}(2011)]%
        {Who2011}
\bibfield{author}{\bibinfo{person}{{World Health Organization}}.}
  \bibinfo{year}{2011}\natexlab{}.
\newblock \bibinfo{title}{World report on disability 2011}.
\newblock
\newblock


\end{thebibliography}


\end{document}